\documentclass[twocolumn,10pt,cleanfoot]{asme2e}
\usepackage[dvips]{graphicx}
\usepackage{psfrag}
\usepackage{epsfig}
\usepackage{array}
\usepackage{amssymb}
\usepackage{subeqn}
\usepackage{setspace}

\confshortname{IDETC/CIE 2009}
\conffullname{the ASME 2009 International Design Engineering Technical Conferences \&\\
              Computers and Information in Engineering Conference}

\confdate{August 30-September 2}
\confyear{2009}
\confcity{San Diego}
\confcountry{USA}

\papernum{DETC2009/MR-86328}

\title{The eel-like robot}
\author{F. Boyer$^{1,2}$, D. Chablat$^{1}$, P. Lemoine$^{1,3}$, P. Wenger$^{1}$
    \affiliation{
    $^1$ Institut de Recherche en Communications et Cybern\'etique de Nantes\\
    UMR CNRS n$^\circ$ 6597, 1 rue de la No\"e, 44321 Nantes, France\\
    $^2$ Ecole des Mines de Nantes, 4 rue Alfred-Kastler, Nantes 44307, France \\
    $^3$ Ecole Centrale de Nantes, 1 rue de la No\"e, 44321 Nantes, France\\
    Email: \{Frederic.Boyer, Damien.Chablat, Philippe.Lemoine, Philippe.Wenger\}@irccyn.ec-nantes.fr}
    }

\begin{document}
\maketitle
\begin{abstract}
The aim of this project is to design, study and build an ``eel-like robot'' prototype able to swim in three dimensions. The study is based on the analysis of eel swimming and results in the realization of a prototype with 12 vertebrae, a skin and a head with two fins. To reach these objectives, a multidisciplinary group of teams and laboratories has been formed in the framework of two French projects. 
\end{abstract}
\section*{INTRODUCTION}

As compared with our technological achievements, the performances of fish make the human dream. Among them include their prodigious capacity for acceleration up to 20 times gravity, their speed exceeding 70 km/h, their extraordinary maneuverability: $180^{\circ}$ turn without slowing down and radii of curvature about 10 times their length, while the current vehicles must slow down by half and take radius of about 10 times their length. Efficiency is about 10 times higher than our best submarines, etc... These features alone justify current efforts to understand and reproduce solutions used by the fish in our robotic systems. In this area, under the biomimetics, the first difficulties encountered is described as follows: ``Replicate the performance of a fish by simple imitation of its form and function would be impossible because the development of a vehicle bending so smooth and continuing is beyond the current possibilities of robotics.'' \cite{Triantafyllou:1995}.
Thus, the continuing fish is an essential difficulty of research in this area. The purpose of this project is to enhance biomimetics by producing a prototype eel-like robot that is ``more continuous'' than its counterparts today. The mechanical architecture of the prototype consists in stacked parallel modules sheathed by a continuous flexible part playing the role of the skin. 
\section*{THE EEL SWIMMING}

The object of the biomimetic robotics is to mimic life, to imitate biological systems or to conceive new technologies drawn from the lesson of their study \cite{Cham:2002}. 

In nature, there are two main types of fish, each being subservient to a type of swimming. The first were carangid swimming, as jacks, horse mackerel or pompano \cite{Cham:2002}. The latter are anguilliform such as the moray or the eel whose handling capacity reached records. It is this second type of swim that our project is to achieve. An anguilliform swimmer propels itself forward by propagating waves of curvature backward along its body \cite{Triantafyllou:1995}. In this case, the maneuverability is the result of the high redundancy (hyper-redundant) induced by the deformation of the fish body on the dimensions of the task. Before any investigative technique, the project started with a literature review of biomimetic fish in general and eel in particular. 

The designers studied the system skeleton - muscles - tendons. For the control, we studied the biomimetics of swimming under the ``fluid mechanics'' or more generally, as the allure of swim \cite{Leroyer:2005}. Finally, data on experimental zoologists style swim eels are limited to the planar motion and take the form of films. We can extract displacement and orientation laws of vertebrae as well as bending inter-vertebral taking in our terminology, the sense of curvature \cite{Meunier:2006}. 

From these data we are committed to characterize and identify swimming simplified underlying parameters of minimum distorted set. Thus, we have updated the laws of wave propagation sine curve progressive or retrograde combined with bends (pitch and yaw) \cite{Boyer:2006}. 

To compensate coupling induced by bending, twisting laws are under study. Each vertebra includes 3 degrees of freedom, bending around two planes (yaw / pitch) and twisting around its column. The prototype includes 12 vertebrae (36 dof), a rigid head and a passive and flexible tail. The head is equipped with side wings mimicking the pectoral fins dedicated to control animal roll and pitch.

\section*{PROTOTYPE DESIGN}

\subsection*{Introduction}

Since the beginning of robotics, engineers have constantly adapted their current design technology. For robots, when it comes to technology, is means mainly technology actuators, computers or materials. Thus, browsing history catalogs robots, we find that for the same robot morphology increasing the power of electric motors and their miniaturization has first resulted in the replacement of hydraulic motors, then simplified and reduced the number of parts by removing parallelograms or balance weights. In another vein, increasing the power of computers has helped to devise more complex mechanical structures and the integration of dynamic models. Thus, from simple mechanical architectures such as Cartesian robots or anthropomorphic, new mechanical architectures appear, called parallel mechanism as the Gough-Stewart platform \cite{Merlet:2000}.

For the design of an eel-like robot, we must adapt our constraints in mature technologies. Indeed, we could have used  shape memory of piezoelectric actuators if they had owned the dynamics and power requirements for moving a robot in the water. Also, we have chosen micro-motors with direct current that has the main advantage of being commanded in couple. In a similar vein, we can choose technologies that were unsuitable for mass production for our prototype unit.

\subsection*{Choice of the mechanical architecture}

\begin{figure}[ht]
    \center
    \includegraphics[width=0.8\columnwidth]{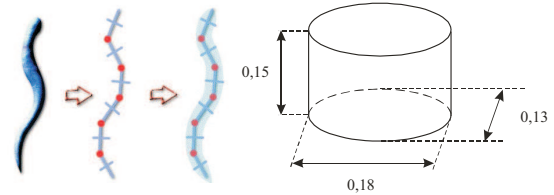}
    \caption{DECOMPOSITION OF THE BODY OF THE EEL VERTEBRA WITH THEIR DIMENSIONS [m]}
    \label{fig:Figure1}
\end{figure}

From the study of biomimetics, it was decided to build the prototype by  stacking  12 vertebrae, each with 3 degrees of freedom of rotation. For our study, the following points were considered:
\begin{itemize}
\item - minimize the inter-vertebral space in order to draw up a continuous deformation model (Figure~\ref{fig:Figure1});
\item - maximize the use of the elliptical section on each vertebra (Figure~\ref{fig:Figure1});
\item - balance the placement of the mechanical to ensure hydrostatic balance;
\item - find the most robust mechanisms with regard to the assembly uncertainties.
\end{itemize}

To accommodate the mechanical parts, the computer and the electronics in the body of the eel, we set the dimensions for each vertebra: focal lengths of 0.18m and 0.13m, height of 0.15m. This amounts to build an eel over 2 meters long when one takes into account the head and tail.
On the basis of an observation of muscular fish, one is tempted to use only linear actuators. For small volume however, there are few robust alternatives to the rotary actuators. Indeed, for a translation, most linear actuators use a rotary actuator, coupled with a ball screw. A lot of energy is lost due to friction. In addition, there are two additional drawbacks as the size of the motor and its guidance and the small amplitude of motion.

\begin{figure}[ht]
    \center
    \includegraphics[width=0.8\columnwidth]{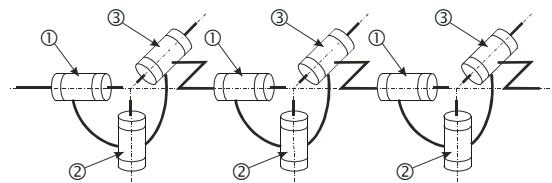}
    \caption{PROTOTYPE OF THE EEL-LIKE ROBOT BASED ON A SERIAL ARCHITECTURE}
    \label{fig:Figure2}
\end{figure}

\begin{figure}[ht]
    \center
    \includegraphics[width=0.8\columnwidth]{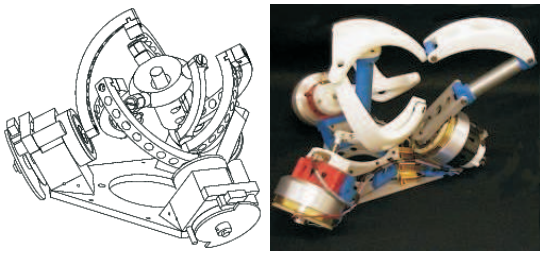}
    \caption{THE AGILE EYE AND SHADE DEVELOPED AT LAVAL UNIVERSITY IN QUEBEC}
    \label{fig:Figure3}
\end{figure}

Similarly, the realization of vertebrae from a serial architecture has been ruled out. Indeed, the use of a serial mechanism as represented in Figure~\ref{fig:Figure2} has the following issues:
\begin{itemize}
\item - the motors location is asymmetric;
\item - the coupling between motor \textcircled{\raisebox{-0.3mm}[0mm][0mm]{1}} and \textcircled{\raisebox{-0.3mm}[0mm][0mm]{3}}  requires a complex assembly for the transfer of constraints between the vertebrae;
\item - the displacement of motor \textcircled{\raisebox{-0.3mm}[0mm][0mm]{2}} causes the displacement of large masses.
\end{itemize}
Finally, we chose a parallel architecture. There are many solutions for a spherical wrist with parallel structure. They are usually classified according to the following properties \cite{Karouia:2003}:
\begin{itemize}
\item - symmetrical / unsymmetrical
\item - isostatic / overconstraint
\item - linear actuators / rotary actuators.
\end{itemize}
However, few technological achievements of ``spherical wrist'' exist today. Among these few prototypes, the best known in robotics is probably the agile eye developed by Clement Gosselin \cite{Gosselin:1994}. This architecture was used to guide a camera in space (hence its name agile eye) or haptic device \cite{Birglen:2002}. It consists in three rotary motors, of which axes intersect at the center of the wrist and three ``legs'' consisting of two revolute joints each, which also cut the main center of the wrist (Figure~\ref{fig:Figure3}). They are the legs making the connection between the fixed part of the mechanism and the camera or the handle of the user.

\begin{figure}[ht]
    \center
    \includegraphics[width=0.8\columnwidth]{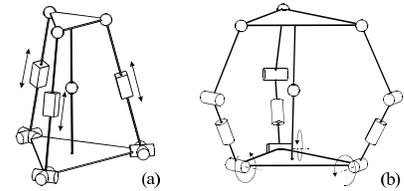}
    \caption{EXAMPLE OF SPHERICAL WRIST WITH THE CENTER OF ROTATION CONSTRAINED BY A PASSIVE BALL-AND-SOCKET JOINT WITH (a) LINEAR ACTUATORS AND (b) ROTARY ACTUATORS}
    \label{fig:Figure4}
\end{figure}

Finally, when one assembles in series such many identical mechanism, all efforts pass through each motor, requiring to strengthen the pivot joints. To eliminate this problem, we studied a family of wrists with a passive ball-and-socket joint in the center of rotation (Figure~\ref{fig:Figure4}).

\begin{figure}[ht]
    \center
    \includegraphics[width=0.999\columnwidth]{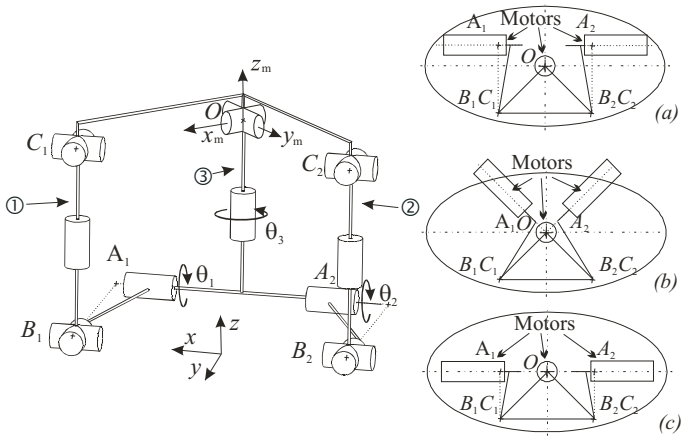}
    \caption{MODELLING VERTEBRAE BASED ON A PARALLEL ARCHITECTURE AND ITS PLACEMENT IN THE ELLIPTICAL ENVELOPE}
    \label{fig:Figure5}
\end{figure}

Thus, an intermediary solution between the agile eye and a spherical parallel wrist with 4 feet can be obtained by actuating the leg with a ball-and-socket joint. Then, a leg is obtained, consisting of a central motorized pivot followed by a universal joint. By affecting the other 2 legs to  the control of both  universal joint rotations, we get the wrist represented in Figure~\ref{fig:Figure5}. This mechanism is derived from the wrist mechanism defined by Agrawal \cite{Agrawal:1995} with rotary actuators instead of linear actuators.

This architecture was chosen in \cite{Chablat:2005}. It is very compact and can be easily located in the elliptical shape of eels. In addition, the kinematics can be reproduced, through the action of the two rods \textcircled{\raisebox{-0.3mm}[0mm][0mm]{1}} and \textcircled{\raisebox{-0.3mm}[0mm][0mm]{2}}, the role of muscles attached to the skeleton and working in addition (in the sense of yaw, for propulsion) and subtraction (in the pitch, for diving). The motor on leg \textcircled{\raisebox{-0.3mm}[0mm][0mm]{3}} allows for the rolling motion. Several motors placements have been testes (Figures~\ref{fig:Figure5} (a), (b) and (c)). The solution (c) has been chosen because in this case, the spindle motors is collinear to the axes of the largest focal length of the ellipse. 

In this case, when the rolling angle is zero, the two coaxial motors operate as a differential gear. Based on this choice, kinematics and geometric models (direct and inverse) of this parallel robot have been developed. During the swimming, only the inverse kinematics is calculated. This model is written as quadratic equations, which can be solved algebraically. This aspect is crucial because of limited processing power of computers shipped. To prevent this actuator from large axial effort, we placed two parallel gears deporting the motors relative to the axis vertebrae. This feature allows to use a motor whose long axis is important.
\subsection*{Assembly of vertebrae}\label{assembly_vertebrae}

\begin{figure}[ht]
    \center
    \includegraphics[width=0.7\columnwidth]{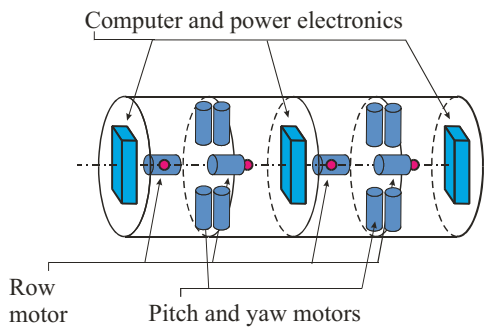}
    \caption{ARRANGEMENT OF THE MECHANICAL, COMPUTER AND POWER ELECTRONICS}
    \label{fig:Figure6}
\end{figure}

The body of the eel is done by mounting vertebrae in serie. In this assembly, the location of actuators, on-board computers and electronics power must be taken into account. After analyzing the needs of on-line calculation, the choice is to allocate a micro-controller to two vertebrae. The motors controlling the pitch and yaw are side by side while the motor controlling the roll is in opposition. This solution is chosen to balance the masses on two vertebrae. Finally, on each section will be set the elliptical skin of the eel-like robot. Figure~\ref{fig:Figure6} represents the position of different elements in the prototype.

\begin{figure}[ht]
    \center
    \includegraphics[width=0.6\columnwidth]{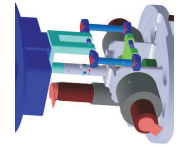}
    \caption{DIGITAL MOCKUP OF A VERTEBRA ON CATIA}
    \label{fig:Figure7}
\end{figure}

Modeled with CATIA, each vertebra is used to simulate the movement of actuators motion and to avoid interferences between the moving parts (Figure~\ref{fig:Figure7}). The joint limits were computed and integrated in the control loop in \cite{Chablat:2008}. 

\begin{figure}[ht]
    \center
    \includegraphics[width=0.8\columnwidth]{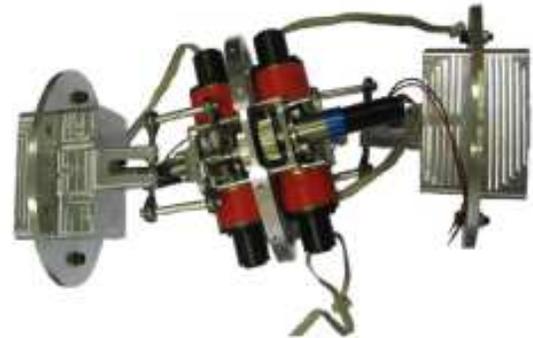}
    \caption{PROTOTYPE OF TWO VERTEBRAE MADE BY TECHNICAL STAFF OF THE IRCCyN}
    \label{fig:Figure7bis}
\end{figure}

\subsection*{Design of the skin}

\begin{figure}[ht]
    \center
    \includegraphics[width=0.99\columnwidth]{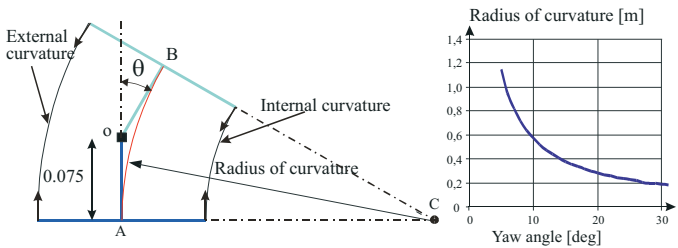}
    \caption{LINK BETWEEN THE RADIUS OF CURVATURE OF EELS AND THE YAW ANGLE}
    \label{fig:Figure8}
\end{figure}

\begin{figure}[ht]
    \center
    \includegraphics[width=0.85\columnwidth]{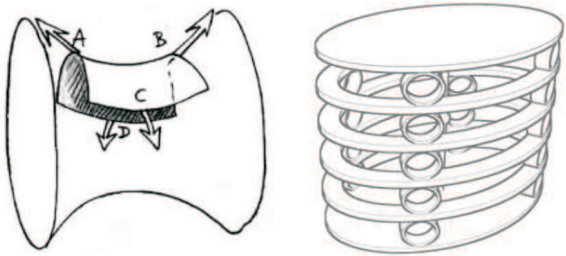}
    \caption{DEFORMATION OF THE SKIN WITHOUT REINFORCEMENT AND WITH THE PROPOSED REINFORCEMENT}
    \label{fig:Figure9}
\end{figure}

The skin of the eel is attached to each vertebra; the objective of our design is to maintain this body to achieve a continuous contact. The difficulty that arises here is the tension between two contradictory goals, which both play a role in the distortion of the body. Indeed, the skin should  offer a very easy distortion in bending-twisting while providing significant resistance to pressure of the fluid. Thus, there are significant distortions that the prototype must undergo. This is illustrated in Figure~\ref{fig:Figure8} on the plane yaw. When the yaw angle is 30 degrees, the external curvature increases by 24\% while the inside curvature decreases by 28\%. To prevent the skin from pleating, a first option would be to precharge the skin when the external curvature is a minimum. Unfortunately, if the skin is made of elastic material (rubber, lycra), its axial stretching is accompanied by an inevitable transverse narrowing  (Figure~\ref{fig:Figure9}).

Also, we have chosen to support such a skin of an underlying structure. For example, this could be based on the stacking of intermediate hollow vertebrae connected by rubber rings (Figure~\ref{fig:Figure9}). Such solution permits to guarantee the continuity in curvature: high deflection, ease of assembly, and its low radial size.
	
The skin is made with three types of materials, plastic rings for reinforcements, chains of rubber to ensure continuity in curvature, and a latex skin to seal and lift between the fluid and the eel-like robot.

\begin{figure}
    \center
    \includegraphics[width=0.85\columnwidth]{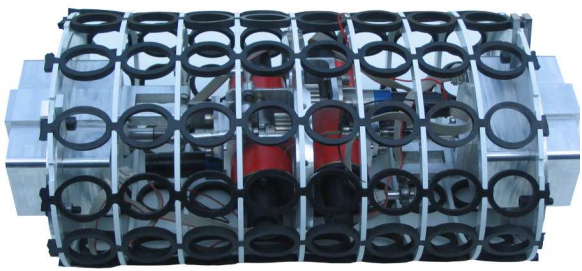}
    \caption{PROTOTYPE OF TWO VERTEBRAE WITH A PART OF THE SKIN WITH RUBBER RINGS AND INTERMEDIATE RIGID SECTIONS}
    \label{fig:Figure10}
\end{figure}
\subsection*{Design of the head}

In the head of the eel-like robot, we have to integrate many sensors (tilting sensor, accelerometers, a measure of relative speed, camera...), a computer as the brain of the fish and two fins to assure the stability. 
The main problem for the integration of the fins is the size of both motors. We need to have collinear axis and we wish to use the same actuators as the vertebrae. 

\begin{figure}
    \center
    \includegraphics[width=0.60\columnwidth]{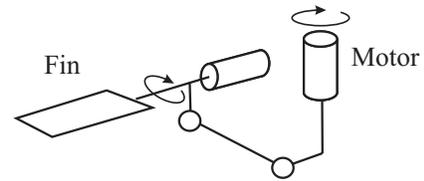}
    \caption{CLOSED-LOOP MECHANISM FOR THE CONTROL OF ONE FIN}
    \label{fig:Figure10bis}
\end{figure}

Two closed-loop mechanisms are used to shift the actuators (Figure~\ref{fig:Figure10bis}). To simplify the design, we used a set of identical parts to the one used for the vertebrae.

\begin{figure}
    \center
    \includegraphics[width=0.85\columnwidth]{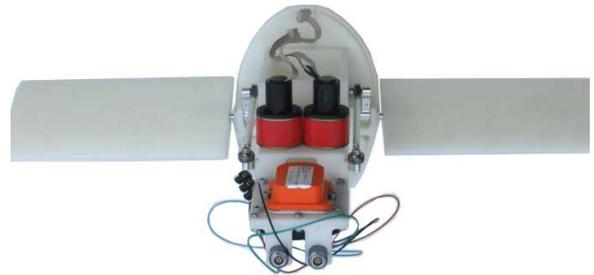}
    \caption{PROTOTYPE OF THE HEAD MADE IN RAPID PROTOTYPING WITH TWO CLOSED-LOOP CHAINS TO CONTROL THE FINS}
    \label{fig:Figure11}
\end{figure}
\begin{figure}
    \center
    \includegraphics[width=0.85\columnwidth]{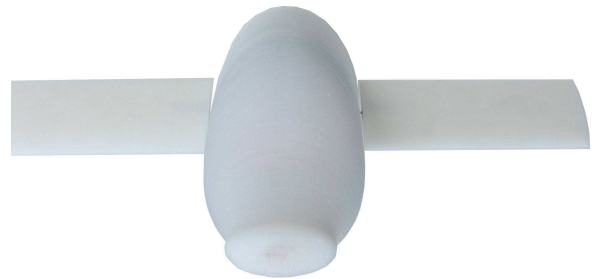}
    \caption{PROTOTYPE OF THE HEAD MADE IN RAPID PROTOTYPING WITH ITS SKIN}
    \label{fig:Figure12}
\end{figure}

For the first prototype, the head is made by rapid prototyping which allows us to test the volume used by the wires (Figure~\ref{fig:Figure11}). The skin of the head is rigid and it is made in rapid prototyping  (Figure~\ref{fig:Figure12}). At the end, it will be covered by a skin made in  latex. 
\section*{SOFTWARE AND HARDWARE IMPLEMENTATION}
\subsection*{Technology constraints}

For this type of project where actuators and control systems are on-board, the definition of space, weight and energy consumption are essential. Another aspect is the wiring system. We must ensure that the various wires for power and data transport do not reduce the manoeuvrability of the eel by introducing couples or additional forces to overcome during the movement. To achieve the deformation of the body of the eel, each joint is powered by three motors located in adjacent vertebrae. The motors are distributed over the whole length of the eel. This architecture provides a natural distribution of mass along the body of the eel, unlike a system actuated by rods or cables with a concentration of actuators in the robot. 
During the design phase, we have to integrate all these concepts to minimize:
\begin{itemize}
\item - the volumes filled by actuators, electronic and drive control to allow a maximum amplitude of deformation of the body of the eel,
\item - the number and the section of the wires in order to obtain a wiring harness that disrupts the least possible movement of deformation,
\item - overall power consumption, so as to ensure a self-sufficient in energy,
\item - overweight related to electrical and electronic components, so as not to penalize the buoyancy of the whole robot.
\end{itemize}

Finally, we have to check that all the components are well distributed to obtain a good balancing of the masses. It is to be noted that the purpose of the swimming process is to guide the head of the eel-like robot. Indeed, it is its position, direction and speed of motion, which are given by the operator. The propulsion for such a motion is generated by the distortion of the eel.

Thus, all information concerning position, direction or progress of the robot are relevant to the head of the eel. It is therefore natural that the head is instrumented with sensors and that it should be taken as reference for the deformations of the rest of the body. Even if some information is not necessary when the robot is tele-operated, they are used to assess the performance of the prototype (mechanical power and rebuild of its trajectory relative to surroundings), making use of a minimum of instrumentation relevant.

As the prototype is tele-operated, it is necessary to establish a communication channel between the operator and the eel. The operator is then a part of the control loop by giving the instructions to adjust the trajectory like an airplane pilot does.

Finally, it is necessary to periodically perform a number of sequentially computations for each link (calculation of the desired joint angles and position control loop for each motor). The calculations to perform in a given period of time depend on the number of joints or vertebrae. This imposes constraints in terms of overall computing power of control system depending on the number of vertebrae.
\subsection*{Solution for the prototype}

With all these considerations we have chosen a solution based on a set of computers distributed on the vertebrae and on the head of the eel (and whose number depends on the number of actuated joints).

With this architecture, the head of the eel has its own computer to manage different sensors, communication with the operator and to act as the  supervisor of the whole robot. Other computers release the supervisor from operations to be conducted periodically to determine the instructions to be sent to each motor and to control the local position loops.

This solution allows at first to establish a true parallel execution and secondly to increase the number of computers when the number of vertebrae grows up. The computing power of the system is automatically adjusted in function of the number of vertebrae.

\begin{figure}[ht]
    \center
    \includegraphics[width=0.85\columnwidth]{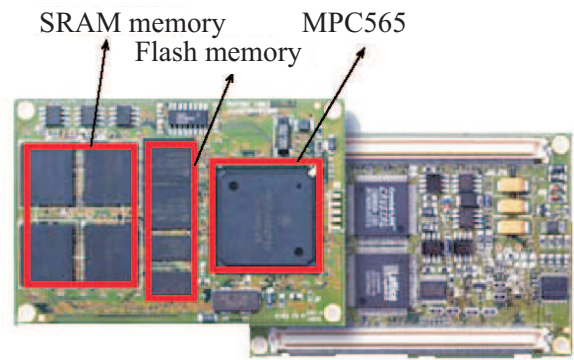}
    \caption{FRONT AND BACK VIEWS OF THE MODULE PHYTEC MPC565}
    \label{fig:Figure21}
\end{figure}

For the material of our application, our choice is on a module of the society Phytec built around the  Motorola microprocessor MPC565 and presented in Figure~\ref{fig:Figure21}. This 32-bit RISC processor of the PowerPC MPC500 family is running at 56 MHz, has a 64-bit floating point unit, and offers opportunities to interface 6 actuators (2 joints) and a set of input/output functions based on the performance of an independent firmware. 

Finally, the MPC565 integrates the management of 3 CAN buses, which can transmit messages of 8 bytes long with a maximum speed of 1Mb/s for each bus. These transmissions are standard asynchronous multi-master and priority management type (the message of the highest priority is not destroyed in case of collision).

Thus, Phytec MPC565 module dissipates a maximum of 2.15W (0.8W for the Motorola processor core), it is only 84x57x3mm (format similar to credit card) and weights only 27g. Of course in our application we only use a reduced set of the module capabilities, thus reducing overall consumption.
The possibility for each MPC565 processor to drive 6 motors allows us to use a single module to manage 2 joints. This option is used as stated in Section \ref{assembly_vertebrae} in alternating motors and electronics/computer bearing vertebrae. 

\begin{figure}
    \center
    \includegraphics[width=0.99\columnwidth]{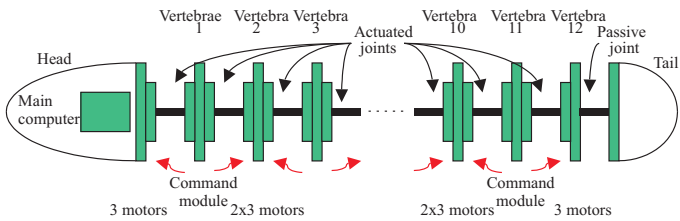}
    \caption{DISTRIBUTION OF THE MODULES ALONG THE EEL}
    \label{fig:Figure22}
\end{figure}

The batteries can be distributed along the eel to balance the masses despite the alternation of the vertebrae. So far, this feature is  not used in the prototype and the volume is replaced by a mass of lead. Figure~\ref{fig:Figure22} shows the distribution of the modules along the body of the eel. Today all the batteries are put in the head thus providing an easy way to exchange them with another set. To obtain a density of 1 the weight of each vertebra must be 2.7Kg and the head 4.5Kg. The total weight of the Lithium-Polymer accumulators is 1.1Kg with a total size of 268x88x72mm.

By bringing together elements by functional modules (actuation and control modules) allows confining  in compact and well defined spaces, reduces wiring interconnection between the computer and electronics (control modules), and helps limit the number of wires needed for electric supply and communication between all the elements. To ensure greater flexibility of the wires, their section is reduced by using for all the modules the same voltage of 37 V (used for motors). Thus reducing the intensity necessary, and converting tension as needed as close as possible to fuel elements with very efficient micro-choppers.

Furthermore, the formation of functional modules creates physical blocks easier to waterproof than a multitude of separate elements.

The application software is based on an OSEK/VDX compatible real-time executive developed at the IRCCyN \cite{Bechennec:2006}. It is a real-time operating system integrating tasks management with fixed priorities, tasks, synchronization, mutual exclusion,  events treatment (interrupts, alarms) and communication services offered by the CAN bus.

Some very specific tasks, such as the management of incremental encoders (actuators position sensors) are managed directly by functions performed by an independent firmware running in real parallelism with respect to the main processor execution. This possibility is offered by some embedded functions that discharge the MPC565 core from many disruptions related to these tasks. The global application is built in a distributed architecture, and consists of a supervisory task, supported by the module located in the head of the eel, and tasks of local control, supported by the modules in the vertebrae. 

However, it is necessary to consider the communication aspect and the synchronization between these various tasks as defined above. Indeed the supervisor task must convey to all the tasks of local control the information corresponding to the current behavior of the eel imposed by the operator. This information relates to the parameters of the undulation to distribute along the body of the eel, and a possible overall curvature, which can be added to the undulation. These data can be transmitted simultaneously to all the tasks of local control through the broadcast on the CAN bus of a single message. With this information, each local task can calculate the joint values with respect to time and location of joints in the chain of the eel. 

The synchronization of the undulation distributed along the body of the eel is provided by this periodic broadcast whose message contains a current time topic.

These functions of synchronization and communication are based on the first CAN bus. The second bus is used for diagnosis and feedback from the vertebrae to the head of the eel. This solution avoids conflicts between control and monitoring/diagnosis messages.
\section*{EXPERIMENTAL RESULTS}

\begin{figure}[hb]
    \center
    \includegraphics[width=0.99\columnwidth]{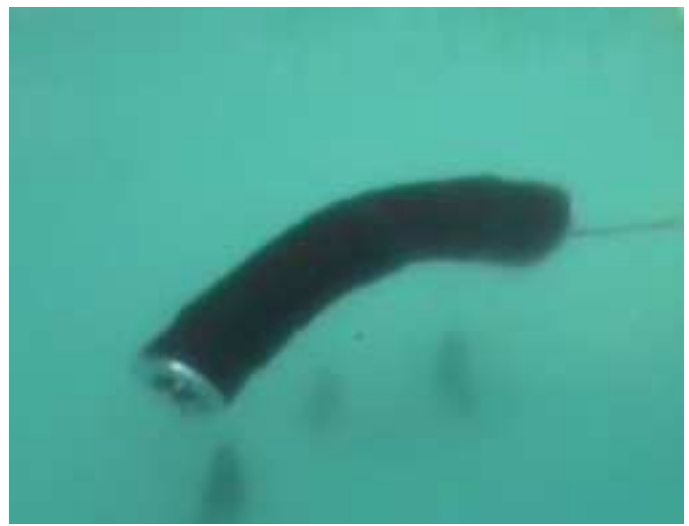}
    \caption{PICTURE OF A 6 VERTEBRAE IN A SWIMMING POOL TO TEST THE WATERTIGHTNESS AND BUOYANCY}
    \label{fig:Figure13}
\end{figure}

The validation of the prototype (i.e. kinematics, electronics and skin) is in progress and applied to a piece of eel (6 vertebrae). 	
These tests aim to validate the watertightness of the skin and the buoyancy of the robot (Figure~\ref{fig:Figure13}).
	
Initial results have validated the watertightness of the skin and showed that the volume of the prototype is very dependent on the differential of the pressure between the inside and outside parts of the robot. So we are testing an autonomous mechanical system based on a calibrated relief valve and a calibrated diving pressure regulator to respectively reduce or increase the inside pressure of the robot with respect to the outside pressure. The purpose of this system is to balance the inside air and outside water pressures to obtain a constant volume of the robot in order to stabilize its buoyancy. The variation of the weight of the embedded air supply during the swimming of the robot should be negligible. This air supply will be a rigid compressed air bottle localized in the head of the eel. Electrical power consumption and capability of body distortion of the eel have also been evaluated and validated during these tests. The theoretical power required for swimming is 15W and the mechanical power delivered by the actuators is 102W, but there was an uncertainty concerning the power consumed by the skin deformation. It was shown that the available mechanical power is sufficient to generate the swimming movement when the eel robot is immersed in water. The capacity of the accumulators is 4Ah and can provide the energy for at least 20 minutes of swimming.

The definition of the tail is not achieved. However, we have decided that a part of the tail will be flexible and it will be made of a part of some rubber, like a flipper. 
\section*{CONCLUSIONS}

From the eel-like robot project, many results have been obtained concerning the mechanical design, which led, on the basis of a serial stacking of parallel modules, to the first version of our prototype. From this, three key issues can be addressed a) the kinematics of the vertebrae, b) the skin, which, even if solutions were proposed, many uncertainties remain, and  c) the buoyancy of the prototype, which will be extended in the future and will lead to systems or ballast bladder.
\section*{ACKNOWLEDGMENTS}

This research was partially supported by the CNRS (``Anguille'' Project) and the ANR project RAAMO (``Robot Anguille Autonome en Milieu Opaque''). The prototype was built by the IRCCyN technical department. F. Brau, M. Canu, G. Branchu, G. Gallot, A. Girin, P. Molina and S. Jolivet are gratefully acknowledged for their technical help for the manufacturing and assembly of the prototype as well as the master students, G. Fadel and P. Lorne.
\bibliographystyle{}

\begin{thebibliography}{99}
\bibitem{Triantafyllou:1995}
Triantafyllou, M.S. and Triantafyllou, G. S., ``An efficient swimming machine,'' Scientific American, pp. 65--70, March 1995.
\bibitem{Cham:2002} 
Cham J.G., Bailey S.A., Clark J.E., Full R.J., Cutkosky M.R., ``Fast and Robust: Hexapedal Robots via Shape Deposition Manufacturing,'' The Int. Journal of Robotics Research, vol. 21, no. 10, pp. 869--882(14), October 2002.
\bibitem{Leroyer:2005} 
Leroyer, A. and Visonneau, M.,  ``Numerical methods for RANSE simulations of a self-propelled fish-like body,'' Journal of Fluids and Structures 20, pp. 975--991, 2005.
\bibitem{Meunier:2006} 
Meunier, F. J. and Ramzu, , M. Y., ``La R\'egionalisation Morphofonctionelle de l'Axe Vert\'ebral chez les T\'el\'eost\'eens en Relation avec le Mode de Nage,'' Compte Rendus de L'Acad\'emie des Sciences, C.R. Palevol 5, 2006.
\bibitem{Boyer:2006} 
Boyer F., Porez, M. and Khalil, W., ``Macro-continuous torque algorithm for a three-dimensional eel-like robot,'' IEEE Robotics transaction, Vol. 22, pp.763--775, 2006
\bibitem{McIsaac:1999}
McIsaac, K.A., Ostrowski, J.P., ``A geometric approach to anguilliform locomotion modelling of an underwater eel robot,'' IEEE Int. Conf. Robot. and Autom., ICRA 1999, pp. 2843--2848.
\bibitem{Merlet:2000}
Merlet, J.P., ``Parallel Robots,'' Kluwer Academic Publishers, 2000. 
\bibitem{Karouia:2003}
Karouia, M., ``Conception structurale de m\'ecanismes parall\'eles sph\`eriques,'' Doctorat thesis, RI 2003-26, \'Ecole Centrale de Paris, 2003.
\bibitem{Gosselin:1994}
Gosselin, C. and Hamel, J.F., ``The agile eye: a high performance three-degree-of-freedom camera-orienting device,'' IEEE Int. conference on Robotics and Automation, pp.781-787. San Diego, 8-13 May, 1994.
\bibitem{Birglen:2002}
Birglen, L., Gosselin, C., Pouliot, N., Monsarrat, B., and Lalibert\'e, T., ``SHaDe, a new 3-dof haptic device,'' IEEE Transactions on Robotics and Automation, Vol. 18, No. 2, pp. 166--175, 2002.
\bibitem{Agrawal:1995}
Agrawal, S.K, Desmier, G. and Li. S., ``Fabrication and analysis of a novel 3 dof parallel wrist mechanism,'' ASME J. of Mechanical Design, 117(2):343--345, June 1995.
\bibitem{Chablat:2005}
Chablat, D. and Wenger, P., ``Design of a Spherical Wrist with Parallel Architecture: Application to Vertebrae of an Eel Robot,'' ICRA05, Barcelona, April 2005. 
\bibitem{Chablat:2008}
Chablat D., ``Kinematic analysis of the vertebra of a eel-like robot,'' ASME Design Engineering Technical Conferences, August 3-6, New York, USA, 2008.
\bibitem{Bechennec:2006}
Bechennec, J.L., Briday, M., Faucou, S. and Trinquet, Y.,
``Trampoline An Open Source Implementation of the OSEK/VDX RTOS,''
11th IEEE International Conference on Emerging Technologies and Factory Automation (ETFA\'06), Prague, September, 2006.
\end{thebibliography}

\end{document}